\documentclass[10pt]{article}

\usepackage[margin=0.82in]{geometry}
\usepackage{graphicx}
\usepackage{amsmath,amssymb}
\usepackage{booktabs}
\usepackage{multirow}
\usepackage{makecell}
\usepackage{tabularx}
\usepackage{xcolor}
\usepackage{tikz}
\usepackage{microtype}
\usepackage{caption}
\usepackage{float}
\usepackage{placeins}
\usepackage[numbers,sort&compress]{natbib}
\usepackage[colorlinks=true,linkcolor=blue!50!black,citecolor=blue!50!black,urlcolor=blue!50!black]{hyperref}

\usetikzlibrary{arrows.meta,positioning,fit,calc}
\title{TextHOI-3D: Text-to-3D Hand-Object Interaction via Discrete Multi-View Generation and Joint Mesh Optimization}

\author{
  Zixiong Hao\\
  Technical University of Munich\\
  School of Computation, Information and Technology\\
  \texttt{zixiong.hao@tum.de}
  \and
  Zhencun Jiang\\
  Tongji University\\
  Shanghai Research Institute for Intelligent Autonomous Systems\\
  \texttt{zhencunjiang@tongji.edu.cn}
}

\date{}

\begin{document}
\maketitle
\vspace{-1.2em}

\begin{abstract}
Text-conditioned 3D generation has progressed rapidly for images and isolated objects, but producing a hand-object mesh remains challenging: the output must preserve language semantics, cross-view consistency, object geometry, articulated hand shape, and physically plausible contact. We present TextHOI-3D, a staged framework that uses generated multi-view observations as an explicit interface between text-conditioned visual generation and geometry-aware hand-object recovery. TextHOI-3D learns a compact VQ token space for fixed-camera hand-object observations, predicts multi-view visual tokens from text with a CLIP-conditioned visual autoregressive model, and recovers a unified hand-object mesh through prior initialization, multi-view joint optimization, and anti-penetration refinement. The design separates semantic generation from geometric recovery while keeping both stages connected by a discrete multi-view representation. On HO3D-derived evaluations, the multi-view setting reduces object CD from 17.26 mm to 4.92 mm and penetration volume from 5.3721 cm$^3$ to 0.2193 cm$^3$ compared with a single-view counterpart, while improving hand errors and surface F-scores. These results support multi-view visual tokens as an effective intermediate representation for text-driven 3D hand-object mesh creation.
\end{abstract}

\vspace{-0.6em}
\section{Introduction}

Generating a hand interacting with an object is more constrained than generating an isolated object or a single image. A plausible result must satisfy several coupled requirements: the prompt should determine the object category and interaction semantics, all observations should describe the same scene, the object should have stable geometry, the hand should remain anatomically valid, and the final hand-object relation should avoid severe floating or penetration. These requirements are difficult to satisfy with a direct text-to-mesh formulation, especially because hand-object contact regions are frequently occluded and interaction data with paired text and meshes are limited.

We study a staged alternative: first generate coherent multi-view visual evidence from text, then recover a unified 3D hand-object mesh. Multi-view images are easier to synthesize than meshes, yet they retain silhouettes, occlusion patterns, and contact cues that can drive 3D recovery. The key question is how to make this intermediate representation compact enough for generation, controllable by language, and useful for geometry.

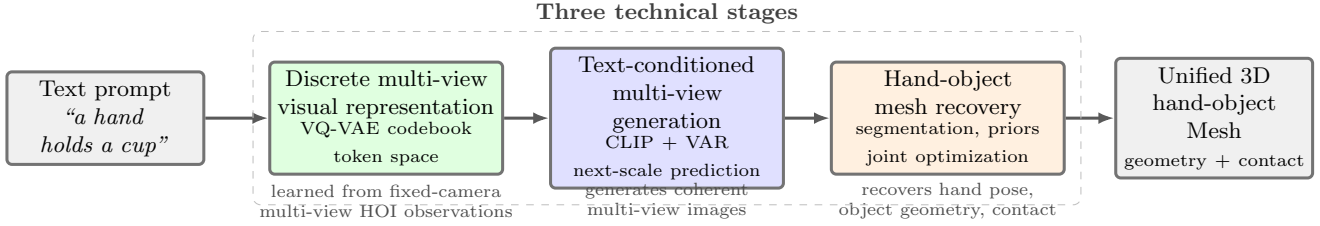
\begin{figure}[H]
  \centering
  \resizebox{\linewidth}{!}{\begin{tikzpicture}[
  font=\small,
  box/.style={draw=black!55, rounded corners=2pt, very thick, align=center, minimum height=1.05cm},
  stage/.style={box, fill=#1!12, text width=3.0cm},
  io/.style={box, fill=gray!12, text width=2.5cm},
  arrow/.style={-{Latex[length=2.2mm]}, very thick, draw=black!65},
  note/.style={font=\scriptsize, align=center, text=black!70},
  group/.style={draw=black!25, rounded corners=3pt, dashed, inner sep=6pt}
]
\node[io] (text) at (0,0) {Text prompt\\ \textit{``a hand holds a cup''}};
\node[stage=green] (repr) at (3.9,0) {Discrete multi-view\\visual representation\\ \scriptsize VQ-VAE codebook\\token space};
\node[stage=blue] (gen) at (7.8,0) {Text-conditioned\\multi-view generation\\ \scriptsize CLIP + VAR\\next-scale prediction};
\node[stage=orange] (rec) at (11.7,0) {Hand-object\\mesh recovery\\ \scriptsize segmentation, priors\\joint optimization};
\node[io] (mesh) at (15.4,0) {Unified 3D\\hand-object Mesh\\ \scriptsize geometry + contact};

\draw[arrow] (text) -- (repr);
\draw[arrow] (repr) -- (gen);
\draw[arrow] (gen) -- (rec);
\draw[arrow] (rec) -- (mesh);

\node[note] at (3.9,-1.15) {learned from fixed-camera\\multi-view HOI observations};
\node[note] at (7.8,-1.15) {generates coherent\\multi-view images};
\node[note] at (11.7,-1.15) {recovers hand pose,\\object geometry, contact};

\node[group, fit=(repr) (gen) (rec)] (system) {};
\node[font=\bfseries\small, text=black!75] at (system.north) [yshift=0.25cm] {Three technical stages};
\end{tikzpicture}}
  \caption{TextHOI-3D overview. The system maps a text prompt to a unified 3D hand-object mesh through three technical stages: discrete multi-view representation learning, text-conditioned multi-view generation, and joint mesh recovery.}
  \label{fig:system_overview}
\end{figure}

We address this challenge with TextHOI-3D, shown in Fig.~\ref{fig:system_overview}. First, a VQ-VAE learns a discrete token space over fixed-camera multi-view hand-object observations. Instead of encoding each view independently, the observations of the same scene are organized as a single stacked tensor, enabling one encoder and codebook to learn scene-level visual tokens. Second, a text-conditioned visual autoregressive model predicts the token maps in a coarse-to-fine manner. CLIP text features~\cite{radford2021learning} are injected through global AdaLN modulation and local cross-attention, allowing the model to control both global interaction semantics and local object attributes. Third, generated views are converted into a 3D hand-object mesh through a multi-view recovery pipeline that combines segmentation, inpainting, object and hand priors, joint optimization, and anti-penetration refinement.

Our contributions are:
\begin{itemize}
  \item A staged text-to-multi-view-to-mesh formulation for text-driven 3D hand-object interaction, instantiated as TextHOI-3D.
  \item A discrete multi-view representation that compresses fixed-camera hand-object observations into VQ tokens suitable for autoregressive generation.
  \item A CLIP-conditioned next-scale generation module that combines global AdaLN modulation and token-level cross-attention for text-controlled multi-view synthesis.
  \item A recovery pipeline that combines segmentation, inpainting, object and hand priors, multi-view optimization, and anti-penetration refinement into a unified hand-object mesh output.
\end{itemize}

\section{Related Work}

\paragraph{Text-conditioned 3D generation.}
Recent text-to-3D methods use optimization, diffusion priors, or multi-view generation to connect language with 3D content. DreamFusion~\cite{poole2022dreamfusion} optimizes 3D representations with score distillation, while Zero123~\cite{liu2023zero}, MVDream~\cite{shi2023mvdream}, and MVDiffusion~\cite{tang2023MVDiffusion} emphasize novel-view or multi-view image generation. Text2HOI~\cite{cha2024text2hoi} generates text-guided 3D hand-object motion conditioned on a canonical object mesh. In contrast, TextHOI-3D targets text-to-mesh hand-object creation through generated multi-view observations and a static mesh recovery stage. This formulation is complementary to motion generation: it focuses on recovering object geometry, articulated hand shape, and local contact from visual evidence rather than producing a temporal interaction sequence.

\paragraph{Hand-object reconstruction.}
Hand-object reconstruction has been studied with paired hand and object meshes, contact priors, signed-distance constraints, and optimization-based alignment~\cite{hasson2019learning,hasson2020leveraging,cao2021rho,ye2022what,chen2022alignsdf,chen2023gsdf,zhang2024moho}. Parametric hand models such as MANO~\cite{romero2017embodied} and recent hand estimators such as HaMeR~\cite{pavlakos2024hamer} provide strong hand priors, while large reconstruction models such as LRM~\cite{hong2023lrm} and InstantMesh~\cite{xu2024instantmesh} improve sparse-view object initialization. TextHOI-3D uses these priors as initialization and constraints, then resolves hand-object alignment with multi-view observations and interaction losses.

\paragraph{Discrete visual tokens and autoregressive generation.}
VQ-VAE~\cite{van2017neural}, VQ-VAE-2~\cite{razavi2019generating}, and VQGAN~\cite{esser2021taming} convert visual data into discrete latent codes, enabling transformer-based image generation. Pixel-level autoregression~\cite{van2016pixel,chen2020generative} models long raster sequences, while VAR~\cite{tian2024visual} predicts token maps scale by scale, shortening the generation path and preserving spatial structure. We use this coarse-to-fine paradigm for multi-view hand-object tokens, where each token implicitly represents a scene-level multi-view unit.

\section{Method}

\subsection{System Overview}

Given a text prompt $y$, TextHOI-3D outputs a hand mesh $M_h$, an object mesh $M_o$, and their relative spatial relation. The pipeline is staged. A discrete representation module maps fixed-camera multi-view observations into token maps. A text-conditioned generator predicts such token maps from $y$. A recovery module then converts generated views into a unified mesh pair. This design keeps the text generation problem in a compact visual token space while leaving geometric consistency and interaction plausibility to the final 3D optimization stage.

\subsection{Discrete Multi-View Visual Representation}

For each hand-object scene, we render a set of fixed-camera RGB views $\mathcal{I}=\{I_i\}_{i=1}^{N}$, where each $I_i\in\mathbb{R}^{H\times W\times 3}$. Instead of processing views independently, we concatenate them along the channel dimension:
\begin{equation}
  X=\mathrm{Concat}(I_1,\ldots,I_N)\in\mathbb{R}^{H\times W\times 3N}.
  \label{eq:stacked}
\end{equation}
This tensor is not assumed to be pixel-aligned across views. It is a unified organization of the same scene under a fixed camera layout, allowing the encoder to learn statistical cross-view correlations.

\begin{figure}[H]
  \centering
  \includegraphics[width=\linewidth]{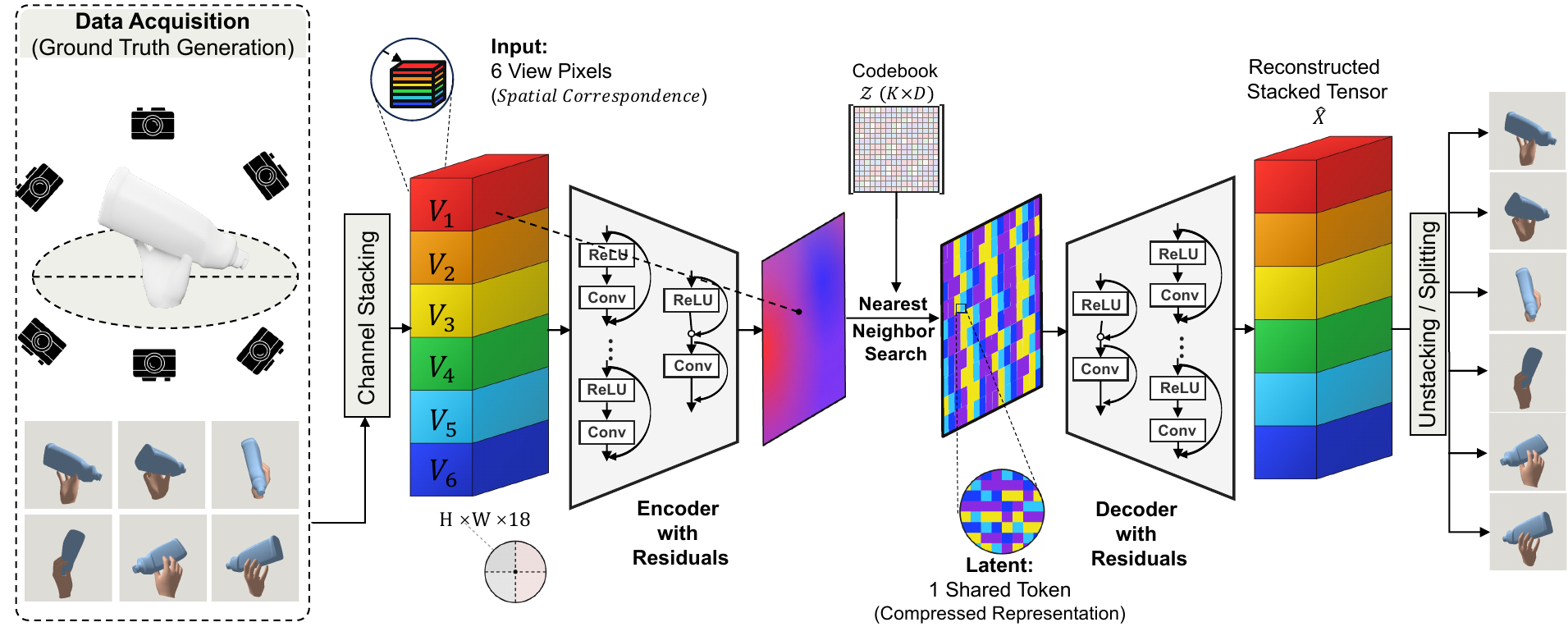}
  \caption{Discrete multi-view representation. Multi-view observations are stacked into a unified tensor, encoded by a residual VQ-VAE, quantized through a shared codebook, and decoded back to synchronized views.}
  \label{fig:discrete_representation}
\end{figure}

The representation model is a VQ-VAE~\cite{van2017neural}. An encoder maps $X$ to a continuous latent feature $z_e=E(X)$; each spatial vector is quantized by a codebook $\mathcal{E}=\{e_k\}_{k=1}^{K}$; a decoder reconstructs the stacked tensor and then splits it back into views. The core data flow is shown in Fig.~\ref{fig:discrete_representation}. The training objective is
\begin{equation}
  \mathcal{L}_{\mathrm{VQ}} =
  \|X-\hat{X}\|_2^2
  + \|\mathrm{sg}[z_e]-z_q\|_2^2
  + \beta\|z_e-\mathrm{sg}[z_q]\|_2^2,
  \label{eq:vq_loss}
\end{equation}
where $z_q$ is the quantized latent, $\hat{X}=G(z_q)$, and $\mathrm{sg}[\cdot]$ denotes stop-gradient. In our implementation, the highest-resolution token map is $16\times16$ with a codebook of 4096 entries. Thus the generator predicts 256 discrete visual tokens instead of dense pixels.

This representation has two roles. It provides reconstruction fidelity for multi-view observations, and it also creates a compact vocabulary for conditional generation. A token therefore corresponds to a multi-view scene unit in the learned latent space, not to an isolated patch from one image.

\subsection{Text-Conditioned Multi-View Generation}

Let $R=(r_1,\ldots,r_K)$ be the multi-scale token pyramid constructed from the VQ latent space. The generator follows the next-scale prediction formulation~\cite{tian2024visual}:
\begin{equation}
  p(R\mid y)=\prod_{k=1}^{K}p(r_k\mid r_{<k},y).
  \label{eq:next_scale}
\end{equation}
At scale $k$, the model predicts the complete token map $r_k$ conditioned on all coarser maps $r_{<k}$ and the text prompt. This differs from raster next-token prediction, where long one-dimensional dependencies can weaken global structure. For hand-object scenes, coarse scales determine the global hand-object layout, while finer scales add silhouettes, local object parts, and contact details.

\begin{figure}[H]
  \centering
  \includegraphics[width=\linewidth]{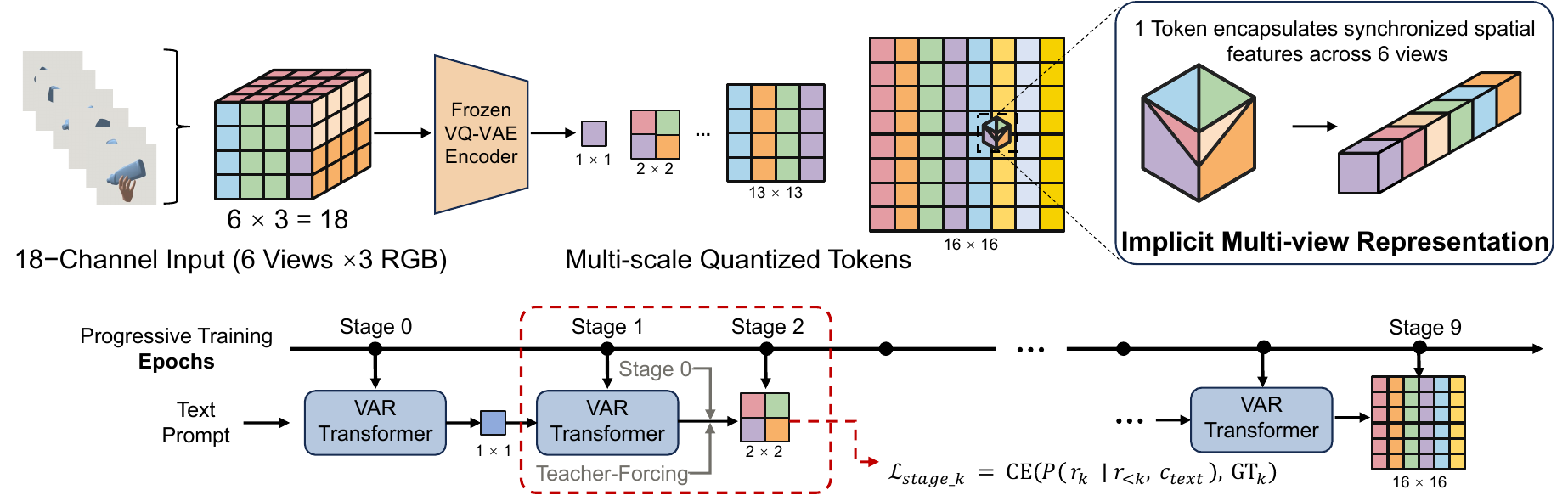}
  \vspace{0.4em}
  \includegraphics[width=\linewidth]{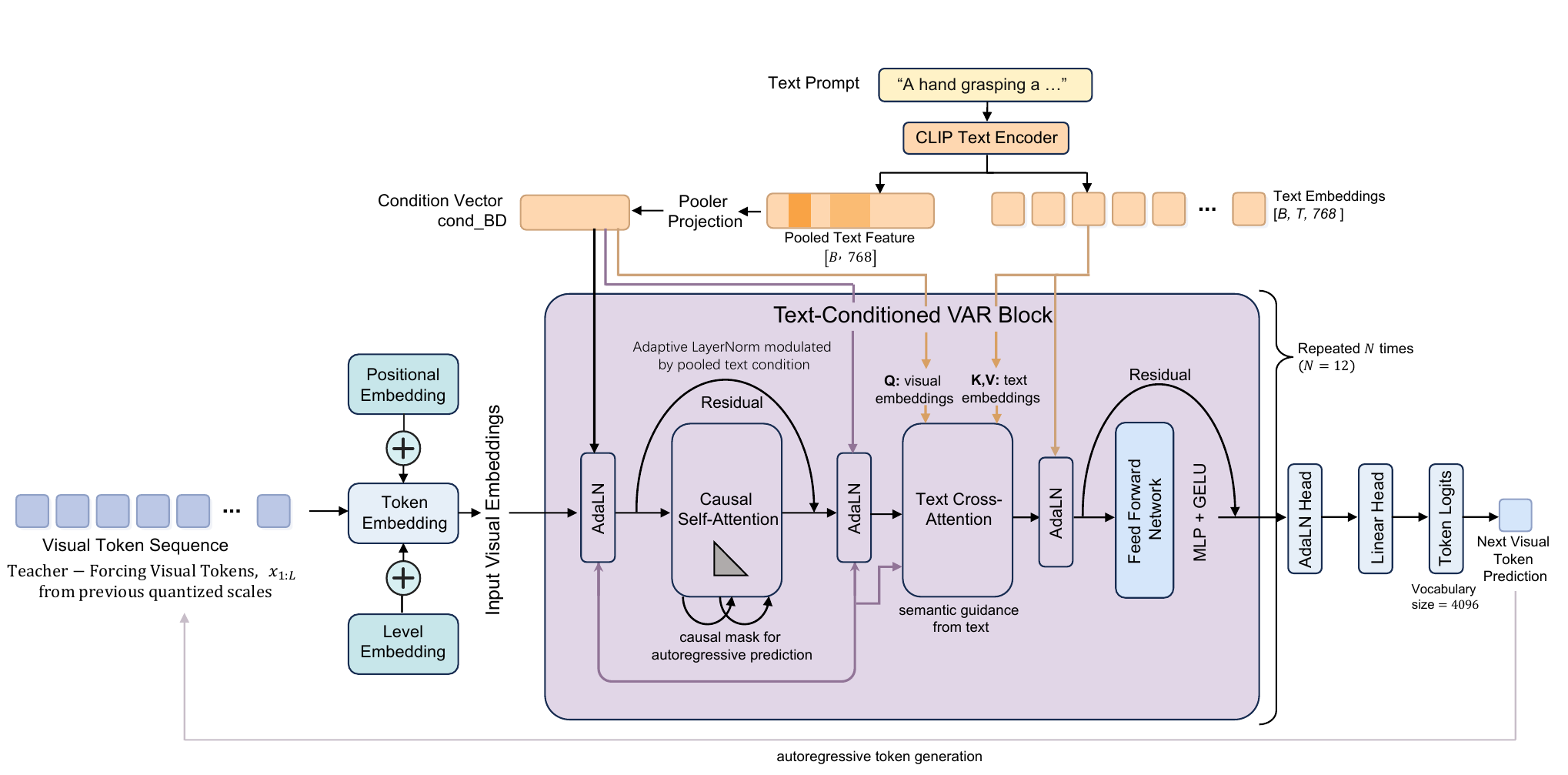}
  \caption{Text-conditioned multi-view generation. The upper diagram shows progressive next-scale token prediction over the discrete multi-view latent space; the lower diagram details the transformer block where CLIP text features provide global AdaLN modulation and local cross-attention guidance.}
  \label{fig:text_var}
\end{figure}

The overall generation process and the internal text-conditioned block are shown in Fig.~\ref{fig:text_var}. A CLIP text encoder extracts a token-level sequence $c_{\mathrm{seq}}$ and a pooled global feature $c_g$. The global feature controls the visual hidden states through AdaLN:
\begin{equation}
  \mathrm{AdaLN}(h,c_g)=\gamma(c_g)\odot\mathrm{LN}(h)+\delta(c_g),
  \label{eq:adaln}
\end{equation}
while token-level text features are injected with cross-attention:
\begin{equation}
  \mathrm{CrossAttn}(H,c_{\mathrm{seq}})=
  \mathrm{Softmax}\!\left(\frac{(HW_Q)(c_{\mathrm{seq}}W_K)^\top}{\sqrt{d}}\right)c_{\mathrm{seq}}W_V.
  \label{eq:cross_attention}
\end{equation}
The former stabilizes global semantics, such as the object class and interaction type; the latter aligns local text attributes, such as handles or grasping cues, with visual token refinement.

Training uses teacher forcing over the token pyramid. At inference time, classifier-free guidance~\cite{ho2022classifier} strengthens text control, and top-$k$/top-$p$ sampling~\cite{holtzman2019curious} balances stability and diversity. The final token map is decoded by the frozen VQ decoder to produce multi-view hand-object images.

\subsection{Hand-Object Mesh Recovery}

\begin{figure}[H]
  \centering
  \resizebox{\linewidth}{!}{\begin{tikzpicture}[
  font=\small,
  box/.style={draw=black!55, rounded corners=2pt, very thick, align=center, minimum height=0.95cm},
  module/.style={box, fill=#1!13, text width=2.45cm},
  arrow/.style={-{Latex[length=2.2mm]}, very thick, draw=black!65},
  note/.style={font=\scriptsize, align=center, text=black!70}
]
\node[module=gray] (mv) at (0,0) {Generated\\multi-view HOI\\images};
\node[module=green] (seg) at (3.0,0) {HSV hand/object\\segmentation};
\node[module=cyan] (inp) at (6.0,0) {SD LoRA\\dual inpainting\\ \scriptsize hand and object};
\node[module=yellow] (init) at (9.0,0.8) {Object branch\\InstantMesh};
\node[module=yellow] (hand) at (9.0,-0.8) {Hand branch\\OmniHands};
\node[module=orange] (repair) at (12.0,0) {Mesh repair\\and SDF filtering};
\node[module=purple] (opt) at (15.0,0) {Multi-view\\joint optimization\\ \scriptsize reprojection, mask, contact};
\node[module=red] (refine) at (18.0,0) {Anti-penetration\\refinement};

\draw[arrow] (mv) -- (seg);
\draw[arrow] (seg) -- (inp);
\draw[arrow] (inp) -- (init);
\draw[arrow] (inp) -- (hand);
\draw[arrow] (init) -- (repair);
\draw[arrow] (hand) -- (repair);
\draw[arrow] (repair) -- (opt);
\draw[arrow] (opt) -- (refine);

\node[note] at (9.0,1.75) {object geometry prior};
\node[note] at (9.0,-1.75) {hand pose and mesh prior};
\node[note] at (15.0,-1.25) {aligns hand, object,\\and interaction in one frame};
\end{tikzpicture}}
  \caption{Hand-object mesh recovery. Generated views are segmented and inpainted, then object and hand priors are initialized separately and refined by multi-view joint optimization.}
  \label{fig:mesh_recovery_pipeline}
\end{figure}
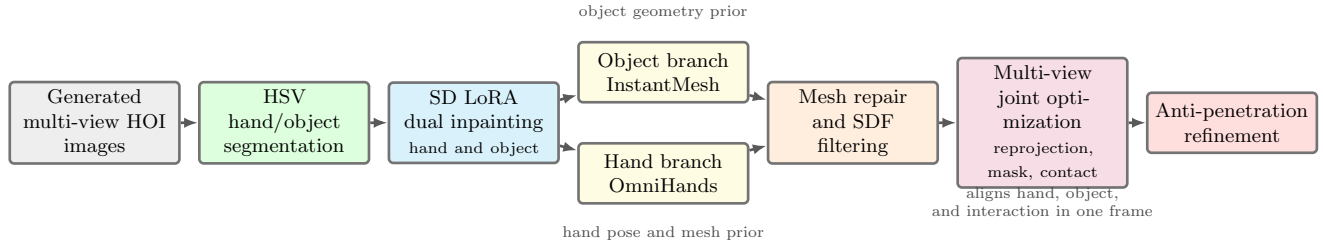

The recovery stage converts generated multi-view images into a unified hand-object mesh. Our pipeline is built around prior initialization and joint refinement rather than direct end-to-end mesh prediction. Fig.~\ref{fig:mesh_recovery_pipeline} shows the main data flow.

We first extract hand and object masks using HSV segmentation, exploiting the stable color distribution of the rendered/synthetic observations. A Stable Diffusion LoRA inpainting module, adapted from latent diffusion models~\cite{rombach2022high}, performs two complementary completion tasks: recovering hand regions occluded by objects and object regions occluded by hands. The object branch sends completed object views to InstantMesh~\cite{xu2024instantmesh} for initial object reconstruction; the hand branch uses OmniHands-style hand priors together with parametric hand modeling~\cite{romero2017embodied,pavlakos2024hamer}. Mesh repair and SDF filtering remove unstable geometry before optimization.

Let the optimization variables be
\[
  \Xi=\{\Theta_h,T_h,R_o,t_o,\Delta M_o\},
\]
where $\Theta_h$ is the hand parameter, $T_h$ is the hand transform, $(R_o,t_o)$ is the object pose, and $\Delta M_o$ is a lightweight object correction. The joint recovery objective is
\begin{equation}
  \Xi^*=\arg\min_{\Xi}
  \lambda_{\mathrm{mv}}\mathcal{L}_{\mathrm{mv}}
  +\lambda_{\mathrm{c}}\mathcal{L}_{\mathrm{contact}}
  +\lambda_{\mathrm{p}}\mathcal{L}_{\mathrm{pen}}
  +\lambda_{\mathrm{r}}\mathcal{L}_{\mathrm{reg}}.
  \label{eq:joint_opt}
\end{equation}
The multi-view term combines reprojection and mask consistency. The contact and penetration terms encourage plausible proximity while discouraging non-physical intersections:
\begin{equation}
  \mathcal{L}_{\mathrm{contact}}=\sum_{v\in\mathcal{C}_h}\max(0,d(v,V_o)-\tau_c),
  \quad
  \mathcal{L}_{\mathrm{pen}}=\sum_{v\in V_h}\max(0,-\phi_o(v)).
  \label{eq:contact_pen}
\end{equation}
Here $\mathcal{C}_h$ is a candidate contact region on the hand, $V_o$ is the object surface, and $\phi_o(\cdot)$ is the object signed distance field. Optimization proceeds in three stages: global multi-view alignment, interaction correction, and local refinement with anti-penetration post-processing.

\section{Experiments}

\subsection{Experimental Setup}

\paragraph{Data and rendering.}
We build experiments from HO3D~\cite{hampali2020honnotate}, a standard hand-object benchmark with YCB-Video objects and paired hand/object geometry. The representation and generation modules use 16,291 rendered frames, split into 14,662 training frames and 1,629 validation frames. Meshes are rendered with PyTorch3D~\cite{ravi2020accelerating} at $256\times256$ resolution under a fixed sparse camera rig. The rig follows a Zero123++-style orbit: azimuth angles start from $30^\circ$ with $60^\circ$ increments, and elevations alternate between $20^\circ$ and $-10^\circ$. For the recovery ablation, we evaluate on nine HO3D-derived examples with ground-truth hand and object meshes. This setting is a controlled input-view ablation rather than a cross-paper leaderboard comparison.

\paragraph{Implementation details.}
The VQ-VAE takes an $H\times W\times18$ stacked tensor as input. Its latent map has spatial size $16\times16$, the codebook contains 4096 entries, each code has dimension 32, and the encoder base width is 160. We train the representation with AdamW~\cite{loshchilov2017decoupled}, learning rate $1\times10^{-4}$, cosine scheduling, 100 epochs, and commitment weight $\beta=0.25$ on three NVIDIA Quadro RTX 6000 GPUs. After VQ training, the encoder and decoder are frozen. The generator is trained with teacher forcing over the next-scale token pyramid; CLIP text features provide global AdaLN modulation and token-level cross-attention. At inference time, classifier-free guidance and top-$k$/top-$p$ sampling are used before decoding the final token map with the frozen VQ decoder. The recovery stage uses HSV hand/object segmentation, Stable Diffusion v1.5 LoRA inpainting, InstantMesh for object initialization, OmniHands-style hand priors, mesh repair, SDF filtering, multi-view joint optimization, and anti-penetration refinement.

\paragraph{Evaluation protocol.}
We report representation quality with PSNR, SSIM~\cite{wang2004image}, and LPIPS~\cite{zhang2018unreasonable}. Object geometry is measured by Chamfer Distance (CD) and F-score at 5 mm and 10 mm after scale-aware ICP alignment; CD and F-score are computed from 10k uniformly sampled surface points. Hand quality is measured by MPJPE over 21 hand joints and MPVPE over 778 MANO vertices after Procrustes alignment. Interaction quality is measured by penetration volume (PV), percentage of penetrating hand vertices (\%PV), mean/max penetration depth, and contact ratio at a distance threshold. PV is estimated by Monte-Carlo sampling with 100k points in the hand-object bounding volume, and \%PV uses a 1 mm inside-object tolerance. For the single-view and multi-view recovery comparison, all weights, reconstruction modules, optimization stages, and evaluation code are fixed; the only variable is the number of input views.

\subsection{Discrete Representation Quality and Efficiency}

\begin{figure}[!t]
  \centering
  \includegraphics[width=\linewidth]{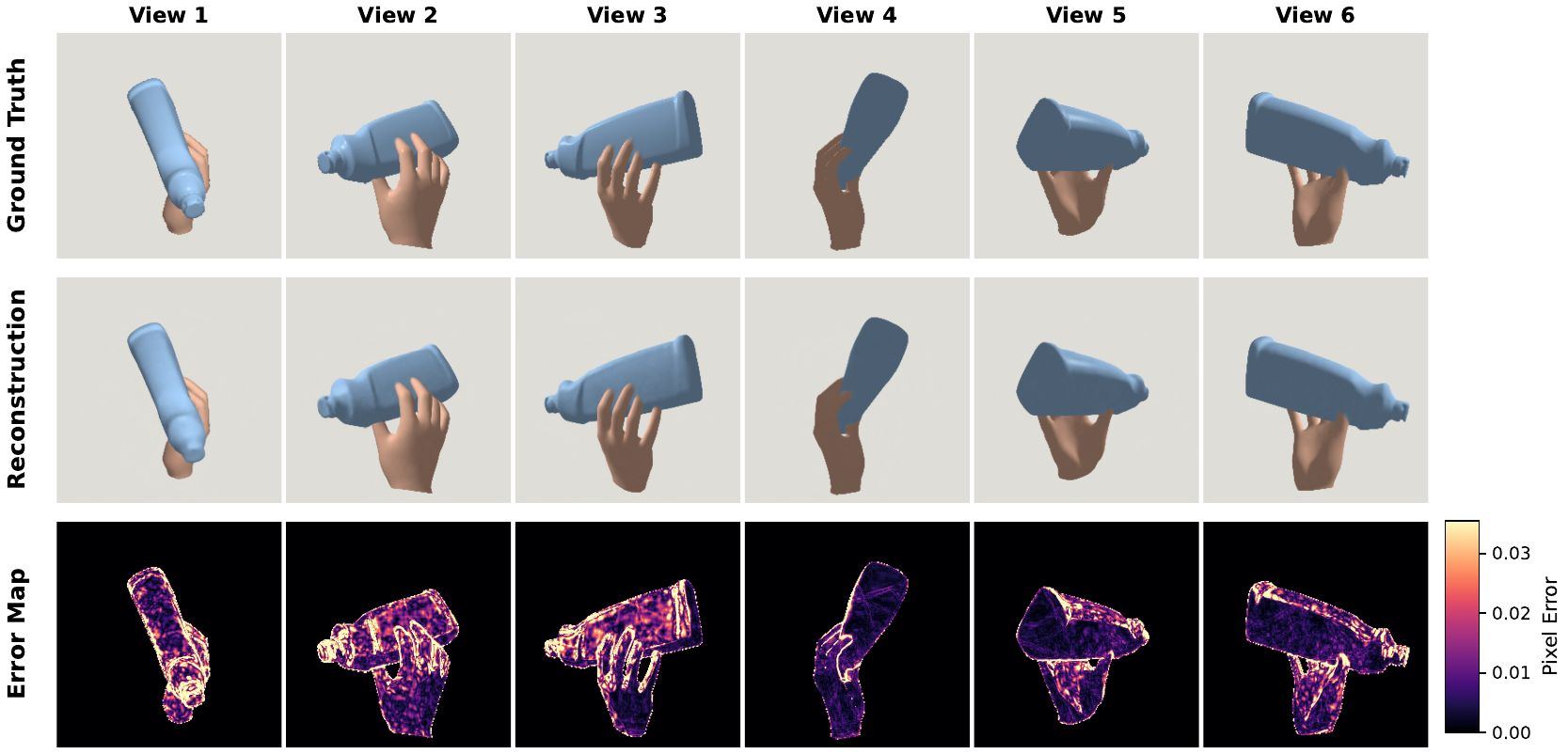}
  \caption{VQ reconstruction examples. The representation reconstructs multi-view hand-object observations while exposing a compact token interface for generation.}
  \label{fig:vq_reconstruction}
\end{figure}

The learned VQ representation preserves multi-view hand-object structure while providing a compact token space for generation. Fig.~\ref{fig:vq_reconstruction} shows representative reconstructions and error maps. The model preserves hand silhouettes, object boundaries, and interaction regions, with most error concentrated near high-frequency edges.

Table~\ref{tab:vq_efficiency} summarizes reconstruction and efficiency. The stacked representation maintains reconstruction quality comparable to a single-view encoder, while processing all views as one scene sample. It also avoids the cost of sequential processing or spatial mosaics: peak memory and step time remain close to the single-view baseline, but view throughput increases substantially.
Efficiency is measured with batch size 1, AMP enabled, 20 warm-up steps, and 50 measured optimization steps on an NVIDIA Quadro RTX 6000 GPU with PyTorch 2.8 and CUDA 12.8.

\begin{table}[!t]
  \centering
  \caption{Compact summary of representation quality and efficiency.}
  \label{tab:vq_efficiency}
  \small
  \setlength{\tabcolsep}{4pt}
  \begin{tabular}{lcccccc}
    \toprule
    Setting & PSNR $\uparrow$ & SSIM $\uparrow$ & LPIPS $\downarrow$ & Mem. $\downarrow$ & Time $\downarrow$ & Throughput $\uparrow$ \\
    & dB & & & GB & ms/step & views/s \\
    \midrule
    Single-view encoder & 42.47 & 0.9834 & 0.0216 & 3.32 & 148.5 & 6.7 \\
    Multi-view stacked VQ & 41.94 & 0.9828 & 0.0225 & 3.33 & 150.1 & 40.0 \\
    Sequential views & -- & -- & -- & 3.74 & 829.1 & 7.2 \\
    Spatial mosaic & -- & -- & -- & 13.29 & 709.4 & 8.5 \\
    \bottomrule
  \end{tabular}
\end{table}

\subsection{Text-Conditioned Multi-View Generation}

\begin{figure}[!t]
  \centering
  \includegraphics[width=\linewidth]{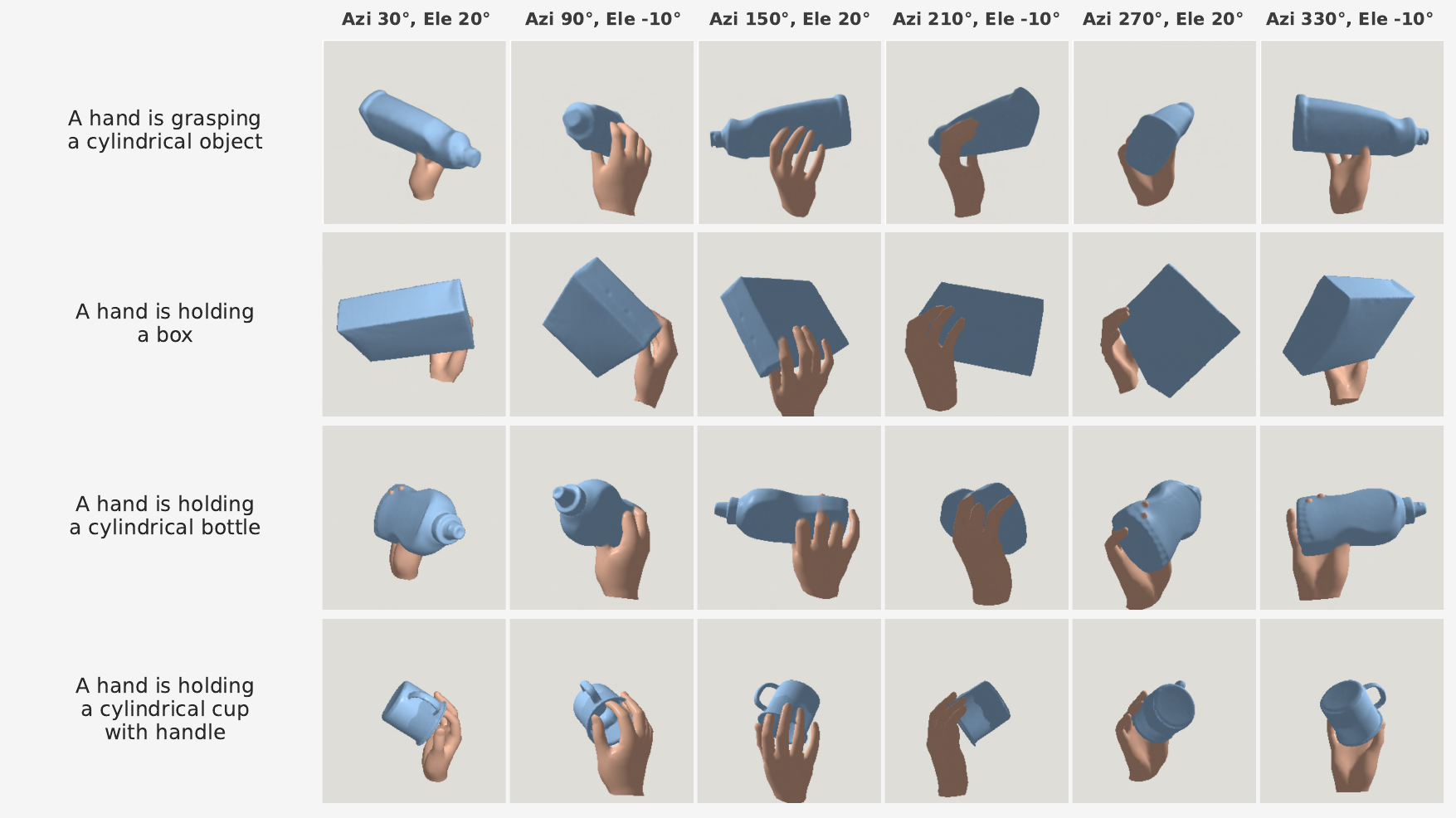}
  \caption{Text-conditioned multi-view generation. The generated views respond to object categories and local structural attributes while preserving cross-view consistency.}
  \label{fig:text_generation}
\end{figure}

Fig.~\ref{fig:text_generation} shows text-conditioned multi-view generation examples. Each row corresponds to one prompt and the generated views share a consistent object structure and hand pose trend. The results indicate that the generated views are not independent images with similar semantics, but visual observations of a shared latent hand-object scene. The coarse-to-fine generation design first establishes global layout and then refines contours and local structures.

\begin{figure}[!t]
  \centering
  \includegraphics[width=\linewidth]{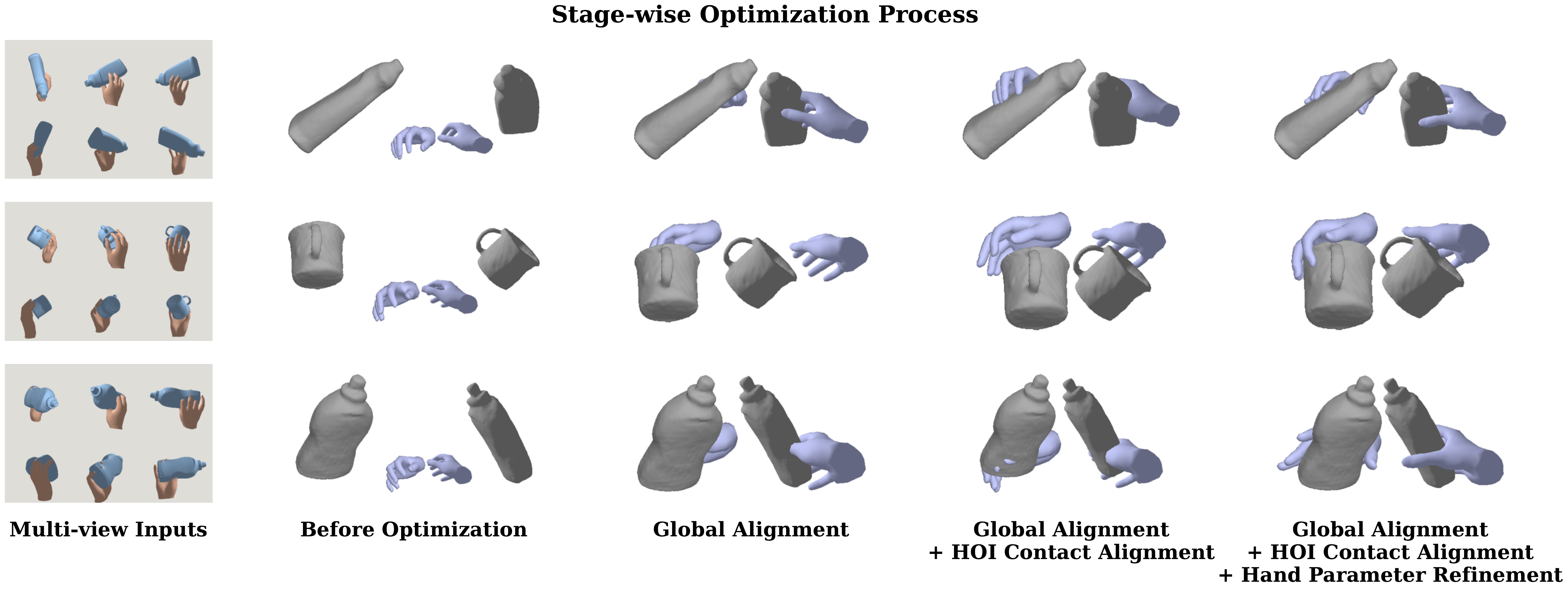}
  \caption{Stage-wise optimization visualization for mesh recovery. The three stages correct global alignment, improve interaction, and refine local hand-object contact.}
  \label{fig:stagewise}
\end{figure}

\subsection{Single-View versus Multi-View Recovery}

The main recovery ablation compares single-view input with multi-view input under the same evaluation protocol. In the single-view setting, object initialization uses Zero123~\cite{liu2023zero} to expand view 0 into sparse views before InstantMesh, while the hand branch uses a single-view hand prior. The multi-view setting directly uses the fixed-camera observations for object initialization, hand priors, and joint optimization. Since the data, weights, recovery modules, and metrics are held fixed, Table~\ref{tab:sv_mv} isolates the effect of multi-view evidence.

\begin{table}[H]
  \centering
  \caption{Single-view versus multi-view recovery. Multi-view observations improve object geometry, hand stability, and penetration suppression.}
  \label{tab:sv_mv}
  \footnotesize
  \setlength{\tabcolsep}{3pt}
  \begin{tabular}{lcccccccc}
    \toprule
    Method & Views & CD $\downarrow$ & F@5 $\uparrow$ & F@10 $\uparrow$ & MPJPE $\downarrow$ & MPVPE $\downarrow$ & PV $\downarrow$ & \%PV $\downarrow$ \\
    & & mm & \% & \% & mm & mm & cm$^3$ & \% \\
    \midrule
    Single-view & 1 & 17.26 & 46.3 & 70.6 & 1.46 & 1.54 & 5.3721 & 1.80 \\
    Multi-view & 6 & \textbf{4.92} & \textbf{92.7} & \textbf{98.6} & \textbf{0.65} & \textbf{0.70} & \textbf{0.2193} & \textbf{0.67} \\
    \bottomrule
  \end{tabular}
\end{table}

The largest gains appear in object geometry: CD decreases from 17.26 mm to 4.92 mm, and F@5/F@10 increase from 46.3/70.6 to 92.7/98.6. This suggests that direct multi-view observations provide more reliable geometry than single-view expansion. Hand errors also decrease, with MPJPE/MPVPE dropping from 1.46/1.54 mm to 0.65/0.70 mm, indicating that multi-view initialization stabilizes the hand prior as well. Finally, PV decreases from 5.3721 cm$^3$ to 0.2193 cm$^3$. This metric is especially important for hand-object recovery because a visually plausible silhouette can still correspond to severe 3D interpenetration.

\begin{figure}[!t]
  \centering
  \includegraphics[width=\linewidth]{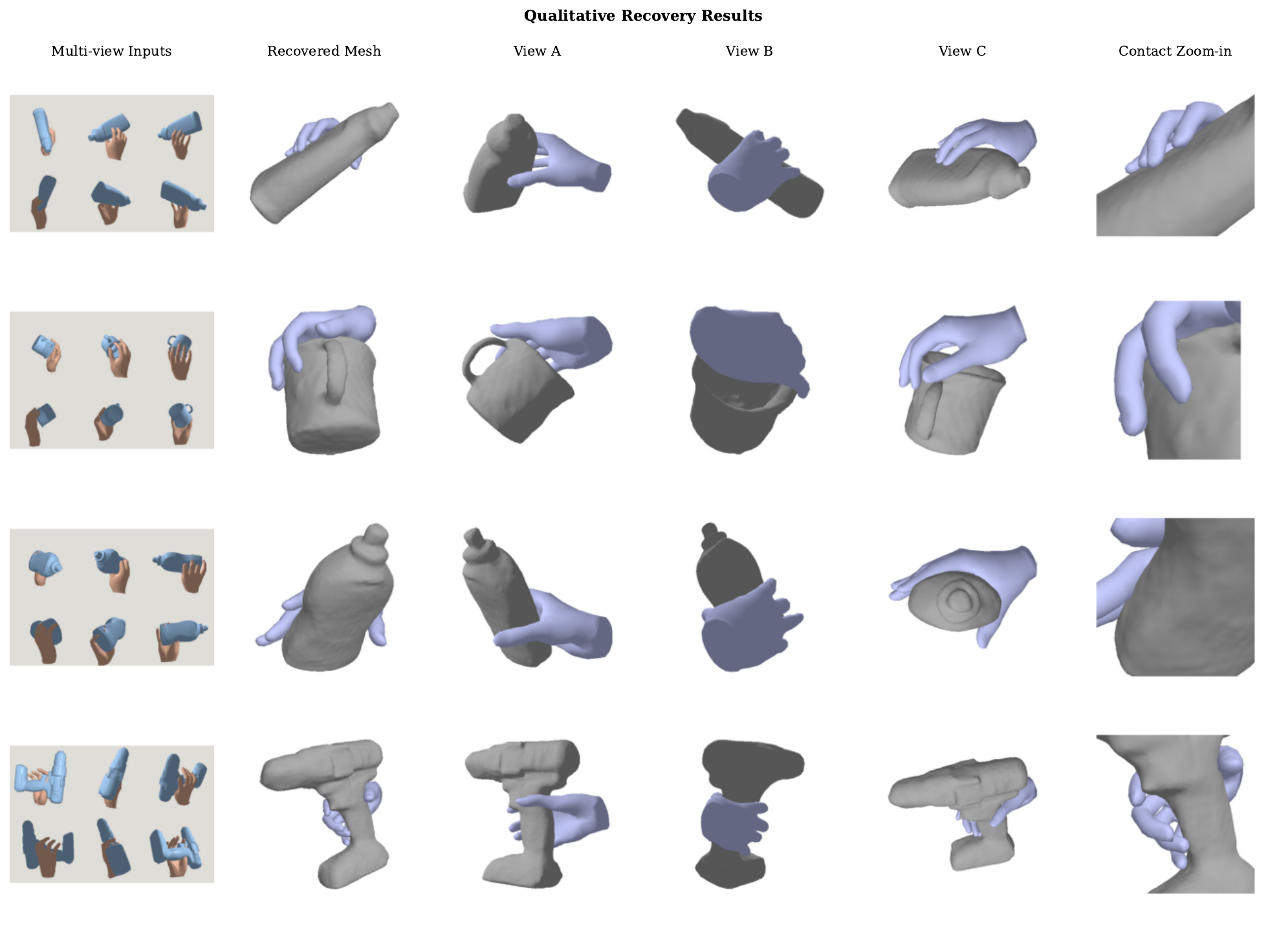}
  \caption{Qualitative mesh recovery results. The recovered hand and object meshes remain coherent under different viewing directions and local contact inspection.}
  \label{fig:mesh_recovery_results}
\end{figure}

\subsection{Interaction Diagnostics}

Beyond aggregate recovery metrics, we report fine-grained interaction diagnostics in Table~\ref{tab:interaction_diagnostics}. MeanPD and MaxPD measure how deep the remaining interpenetration is, while CR@$\tau$ reports the percentage of all 778 MANO vertices within $\tau$ mm of the object surface. These metrics should be interpreted together: PV and penetration depth penalize non-physical overlap, whereas contact ratio checks whether the hand remains close to the object rather than being moved away to avoid intersection.

\begin{table}[H]
  \centering
  \caption{Fine-grained interaction diagnostics for the final multi-view output.}
  \label{tab:interaction_diagnostics}
  \footnotesize
  \setlength{\tabcolsep}{3pt}
  \begin{tabular}{lccccccc}
    \toprule
    Method & PV $\downarrow$ & \%PV $\downarrow$ & MeanPD $\downarrow$ & MaxPD $\downarrow$ & CR@2 $\uparrow$ & CR@5 $\uparrow$ & CR@10 $\uparrow$ \\
    & cm$^3$ & \% & mm & mm & \% & \% & \% \\
    \midrule
    TextHOI-3D & 0.2193 & 0.67 & 4.26 & 6.64 & 0.37 & 1.44 & 2.83 \\
    \bottomrule
  \end{tabular}
\end{table}

The final output has low PV and shallow penetration depth while maintaining non-zero contact ratios. This indicates that the refinement stage does not merely separate the hand from the object; it suppresses severe intersections while preserving local proximity around the interaction region.

\section{Discussion and Limitations}

TextHOI-3D is designed around explicit intermediate interfaces. The discrete visual space compresses multi-view hand-object scenes into a compact representation that is easier to generate than pixels. The text-conditioned VAR model transfers language semantics into coherent multi-view observations. The mesh recovery stage then uses multi-view geometry and interaction constraints to obtain a unified 3D result. This separation makes failures easier to localize: inconsistent generated views mainly affect initialization, while inaccurate segmentation or inpainting mainly affects recovery.

The current evaluation is a controlled system study rather than a broad benchmark submission. Its strength is that the single-view and multi-view variants share the same data, weights, recovery modules, and metrics; its limitation is that the quantitative recovery analysis is restricted to examples with available hand-object ground truth. Future work can close the loop by feeding 3D consistency losses back into the generator and by scaling the evaluation to a larger annotated set.

\section{Conclusion}

We presented TextHOI-3D, a text-to-3D hand-object framework that connects discrete multi-view generation with joint mesh optimization. It learns VQ tokens for hand-object observations, predicts them with a CLIP-conditioned next-scale VAR model, and recovers meshes with multi-view priors, contact constraints, and anti-penetration refinement. The controlled single-view/multi-view ablation confirms that multi-view evidence improves geometry, hand stability, and interaction plausibility.

\FloatBarrier
\bibliographystyle{plainnat}
\bibliography{refs}

\end{document}